\documentclass[conference]{IEEEtran}
\IEEEoverridecommandlockouts
\usepackage{cite}
\usepackage{CJKutf8}
\usepackage{algorithmic}
\usepackage{textcomp}
\def\BibTeX{{\rm B\kern-.05em{\sc i\kern-.025em b}\kern-.08em
    T\kern-.1667em\lower.7ex\hbox{E}\kern-.125emX}}
\usepackage{graphicx}
\usepackage{hyperref}
\usepackage{amsmath,amssymb, amsfonts,empheq,mathtools,mathrsfs,nccmath}
\usepackage{float,subfig, wrapfig}  
\usepackage{arydshln,multirow, multicol, adjustbox}  
\usepackage{relsize} 
\usepackage{color,xcolor} 
\usepackage[linesnumbered,boxed,ruled,commentsnumbered]{algorithm2e}
\begin{document}

\title{A Novel Distributed Representation of News (DRNews) for Stock Market Predictions \thanks{We acknowledge the support by 2016 Jiangsu Science and Technology Programme: Young Scholar Programme (No. BK20160391)}}

\author{\IEEEauthorblockN{1\textsuperscript{st} Ye Ma}
\IEEEauthorblockA{\textit{Department of Financial Mathematics} \\
\textit{School of Science}\\
\textit{Xi'an Jiaotong-Liverpool University}\\
Suzhou, China \\
ye.ma@xjtlu.edu.cn}
\and
\IEEEauthorblockN{2\textsuperscript{nd} Lu Zong}
\IEEEauthorblockA{\textit{Department of Financial Mathematics} \\
\textit{School of Science}\\
\textit{Xi'an Jiaotong-Liverpool University}\\
Suzhou, China \\
lu.zong@xjtlu.edu.cn}
\and
\IEEEauthorblockN{3\textsuperscript{rd} Peiwan Wang}
\IEEEauthorblockA{\textit{Department of Financial Mathematics} \\
\textit{School of Science}\\
\textit{Xi'an Jiaotong-Liverpool University}\\
Suzhou, China \\
Peiwan.Wang@xjtlu.edu.cn}}

\maketitle

\begin{abstract}
In this study, a novel Distributed Representation of News (DRNews) model is developed and applied in deep learning-based stock market predictions. With the merit of integrating contextual information and cross-documental knowledge, the DRNews model creates news vectors that describe both the semantic information and potential linkages among news events through an attributed news network. Two stock market prediction tasks, namely the short-term stock movement prediction and stock crises early warning, are implemented in the framework of the attention-based Long Short Term-Memory (LSTM) network. It is suggested that DRNews substantially enhances the results of both tasks comparing with five baselines of news embedding models. Further, the attention mechanism suggests that short-term stock trend and stock market crises both receive influences from daily news with the former demonstrates more critical responses on the information related to the stock market {\em per se}, whilst the latter draws more concerns on the banking sector and economic policies.
\end{abstract}

\begin{IEEEkeywords}
News network, news embedding, attributed network embedding, Subword model, stock movement prediction, crises warning
\end{IEEEkeywords}

\section{Introduction}\label{sec:introduction}
The stock market, one of the most knowledge-driven place, absorbs information from all channels and releases instant responses through the price oscillation. As the major source of public information, news is considered to be a type of textual data that is voluminous and continuously updated, incorporating vast interactive information between news and various labels from manual annotation or advanced natural language processing (NLP) models. Regarded as one of the driving factors in the stock market, news places major influences on the price evolution in various aspects including the domestic/global economy and financial situations as well as the investors' sentiment. With the aim of facilitating the performance of news-driven stock market predictions, this study proposes a novel Distributed Representation of News (DRNews) model that embeds news as continuous vectors that describe the collective information of its inherited features as well as the inter-textual knowledge across different news articles.

Attention has been drawn to the recent literature on textual representations of news articles and their applications to the deep predictive network of stock movements. For example, the Paragraph Vector model \cite{DBLP:journals/corr/LeM14} is proposed to train news embeddings that facilitate the decision-making process of stock trading strategies \cite{7550882}. Further, Ding et al. (2015) \cite{Ding:2015:DLE:2832415.2832572} suggest to extract event tuples from headlines for news event embedding and associated event-driven stock prediction, whilst Hu et al. (2018) \cite{Hu:2018:LCW:3159652.3159690} use a deep learning framework for news-oriented stock trend prediction by representing news as the average of word vectors. In addition to those unsupervised learning methods, DeepClue \cite{article} represents news in a supervised manner by constructing an end-to-end stock price forecasting model using a neural network as the news encoder and words in the headline as the input. Instead of using solely textual information to predict stock trends, some studies incorporate additional news features, such as news sentiment \cite{e98ec1711a274278b32679ae3f451935,10.1007/978-3-642-24704-0_4,si-etal-2014-exploiting}, public mood \cite{The-effect-of-news-and-public-mood} and news views \cite{Quantifying-Wikipedia-Usage}, into their models for greater reliability and robustness in the predictions. Although the above-mentioned news embedding approaches share the merit that similar news are represented with similar vectors, their news embeddings are limited to the representation of the literal information brought by the news article of interest, rather than the inter-textual knowledge and cross-documental linkages inherited from the news corpus. According to Long's study (2019) \cite{long2019new} that uses the graphic kernel method to predict stock movements based on financial news semantic and structural similarities, it is found that both the content and structure information of news contribute to the prediction with higher weights assigned to the structure information. Moreover, no mechanism has been developed in the existing literature to project additional news features simultaneously into the same vector space of news texts in order to enhance the integrity of the embedding with both contextual and exogenous information.

\begin{figure}[h!]
\centering
\includegraphics[scale=0.6]{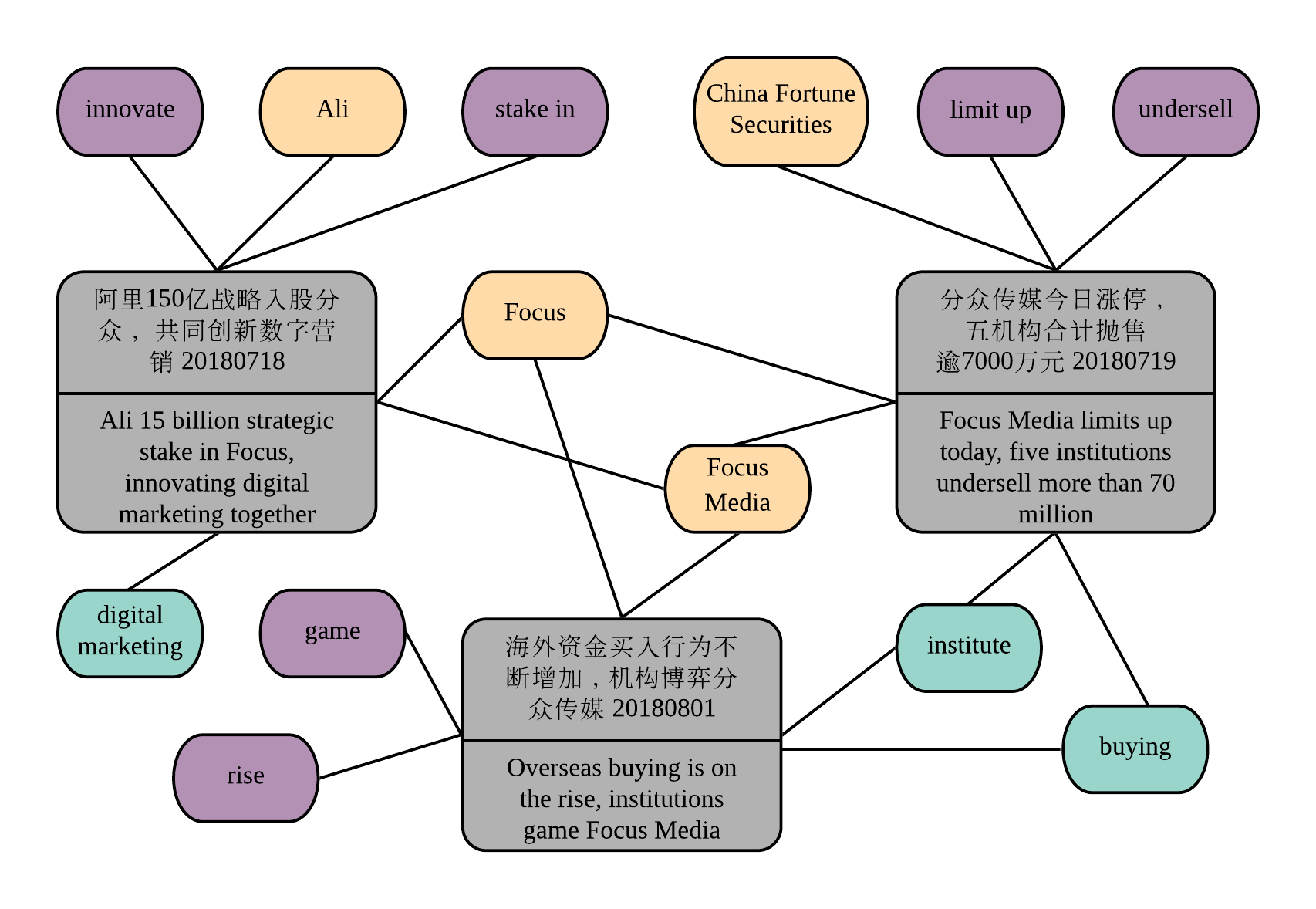}
\caption{A simplified example of DRNews network}
\label{fig:network}
\end{figure}

 In this paper, a Distributed Representation of News (DRNews) model is developed to incorporate the structural linkage between news in addition to a variety of news features. The novelty of DRNews is mainly reflected in two aspects. First, distributed relations between different news are represented in the form of a network so that news articles are embedded with both the semantic and inter-textual knowledge (See Figure \ref{fig:network} for a simplified example of the DRNews network). In particular, different from other established text/sentence embedding models that focus on document-level relationships, such as coherence among sentences \cite{DBLP:journals/corr/HillCK16,D14-1218,DBLP:journals/corr/abs-1803-02893} or the subordination of words to text \cite{DBLP:journals/corr/LeM14}, the proposed DRNews embeddings are trained based on the cross-documental relationship between news stories. According to Figure \ref{fig:network}, the three pieces of news \textbf{\em Ali strategically stakes in Focus with 15 billion, innovating digital marketing together.} \& \textbf{\em Focus Media limits up today with five institutes jointly selling more than 70 million.} \& \textbf{\em Overseas buying rises from institutes' game on Focus Media.} are connected by sharing the same contextual element \textbf{\em Focus Media}. The verbal elements, such as \textbf{\em stake in, limits up, rises}, of the three different news are thus linked through their common nodes and become the latent contextual elements of each other. In this way, the DRNews network describes not only the co-existence of news and its own elements, but also the potential linkage among news through its connected elements. Second, the Subnode model, inspired by the Subword model \cite{2016arXiv160704606B}, is proposed to treat each news node as a bag of node attributes ({\em i.e.} news features including semantic features, category features, time features, {\em etc.}), thus the news vector is computed as the sum of its feature vectors. In the framework of the biased random walk \cite{Grover:2016:NSF:2939672.2939754} and the skip-gram architecture \cite{DBLP:journals/corr/abs-1301-3781}, the Subnode model enriches news embeddings with additional news features, at the same time allowing fast inferences on the embedding of unseen news based on its feature vectors.
 
The major contribution of this study is threefold. First, the DRNews model is developed and applied for two news-driven stock market prediction tasks, i.e. the short-term movement prediction and the stock crises warning, in comparison with other five embedding models. As the results show, DRNews achieves state-of-art performances in both tasks demonstrating its effectiveness in capturing more useful information than the baselines. In particular, a test-set accuracy of $46.24\%$ is achieved in the ternary prediction task of stock movement, whilst a
test-set accuracy of $94.61\%$ and an average of $2.5$-days forewarned period are obtained in the early warning task of stock crises. Second, the Subnode mechanism offers a straightforward solution to enrich node vectors with node attributes and compute embeddings of unseen nodes quickly. Last, the attention-based LSTM (Long Short-Term Memory) \cite{doi:10.1162/neco.1997.9.8.1735} network is adopted as the predictive model, which explicitly identifies the distinguishing dynamics of short-term stock movement and stock crises warning in response to news of different topics and keywords. 

\section{Related work}

\subsection{News embedding}
To represent news articles as vectors, two possible ways are emerged. As news articles are essentially textual data, one could use directly text/sentence embedding models to embed news. Alternatively, news event embedding models are developed with the argument that news stories describe events.

In terms of \textbf{Text/Sentence Embedding}, one of the simplest but robust baseline model \cite{DBLP:journals/corr/abs-1805-01070} is to perform numerical operations on the existing contextual word vectors \cite{DBLP:journals/corr/MikolovSCCD13}. SDAE \cite{Vincent:2010:SDA:1756006.1953039} uses an auto-encoder to project text into a low-dimensional vector space, whilst Paragraph Vector \cite{DBLP:journals/corr/LeM14} learns the distributed representation of sentences and documents by predicting their contexts, {\em i.e.}, words in the sentence/document. Other unsupervised models include Skip-Thought \cite{D14-1218}, FastSent \cite{DBLP:journals/corr/HillCK16} and Quick-Though \cite{DBLP:journals/corr/abs-1803-02893} and learn distributed sentence-level representations regarding to the coherence among sentences across the text. BERT \cite{DBLP:journals/corr/abs-1810-04805} is a powerful pre-trained sentence encoder trained on unsupervised data sets based on the masked language model and next-sentence prediction. As for supervised learning models, most studies use a sentence encoder such as LSTM or Transformer \cite{DBLP:journals/corr/VaswaniSPUJGKP17} to convert word vectors to sentence vectors. The sentence encoder is usually trained on NLP tasks, which could be a single supervised learning task, such as InferSent \cite{DBLP:journals/corr/ConneauKSBB17} which uses only the language inference data, or multiple tasks \cite{DBLP:journals/corr/abs-1803-11175} which involve both unsupervised and labeled corpus. 

On the other hand, \textbf{Event Embedding} models emphasize the mining of event tuples and their projections onto the vector space. Ding et al.'s Event Embedding model \cite{Ding:2015:DLE:2832415.2832572,Ding2016KnowledgeDrivenEE} extracts event tuples $E=(Actor,Predicate,Object)$ from headlines and learn dense vector representations by scoring correct tuples higher than corrupted tuples. Basically, the event embedding model is built on the knowledge graph embedding after decomposing the knowledge graph into tuples of entities, relations, etc. Following Ding et al.'s idea, news events could then be embedded with other knowledge graph embedding models such as TransE \cite{Bordes:2013:TEM:2999792.2999923}, which exploits scoring functions based on the distance between $h+r$ and $t$. In addition to knowledge graph embedding, Event2vec \cite{Setty:2018:ENE:3209978.3210136} employs network embedding to learn distributed representations of summarized news events by proposing an event network where each news event node is connected to the associated entities, the event type and the year of the event. The major limitations of Event2vec are threefold. First, it only works for artificially organized database of already extracted events. Second, the heterogeneous network it constructs raises difficulties in determining the weights of element nodes of different types which leads to further imposed assumptions on the importance of the elements. Third, it lacks a solution to represent unseen events based on the trained network.

In addition to converting news into continuous vectors, structured \textbf{Event Extraction} from news also allows statistical analysis based on the event type \cite{10.1016/j.elerap.2018.02.006,doi:10.1080/14697688.2012.672762}. In the early studies of news data mining, the textual information of news is often regarded as high-dimensional term vectors \cite{azfin,Wang:2012:NTM:2142138.2142443,725072}, which are then considered to lead to a great loss of information \cite{8068217}.

\subsection{Stock market prediction based on deep learning models}

As the central pillar of the financial system, stock market demonstrates great significance as a major channel of capitalization and maintaining liquidity. The applications of deep learning models in stock market predictions have become an emerging research field due to the rapid development and maturation of relevant techniques. With the merit of high feature learning capacity, deep learning enhances the traditional machine learning models and achieves higher level of accuracy as the volume of data increases. 

In the field of stock market prediction, stock trend prediction, return/volatility prediction and stock crisis warning are the three mainstream tasks that attract the most attention. Among the deep learning models that have been adopted in these tasks, classic neural networks, such as long short-term memory (LSTM) \cite{Rather:2015:RNN:2775746.2776067,fischer2018deep,wang2019integrated} and convolutional Neural Networks (CNN) \cite{7550882,Ding:2015:DLE:2832415.2832572} are proven to be the most reliable and widely-adopted techniques handling financial time series \cite{di2016artificial}. Due to recent advances in textual feature engineering which allows the extraction of public events and investor sentiments, integrated predictive methods, such as ensemble learning and attention-based neural networks, have also gained significant popularity in the field. For instance, Chatzis et al. (2018) \cite{chatzis2018forecasting} develop a stacked model combining classification trees, support vector machines, random forests, neural networks, extreme gradient boosting, and deep neural networks to forecast crisis episodes in stock, bond and currency markets. Long et al. (2019) \cite{long2019deep} develop an end-to-end model,  multi-filters neural network (MFNN), to implement feature extraction and price movement prediction. As one of the core techniques in deep learning, the attention mechanism learns the data in the way that allows the machine to assign heavier weights to the more important features in order to improve the forecasting power of the model. Hu et al. (2018) \cite{Hu:2018:LCW:3159652.3159690} apply the Hierarchical Attention Networks \cite{N16-1174} on the stock trend prediction, whilst Vaswani et al. (2017) \cite{8759115} propose the Self-attention Networks, inspired by Transformer \cite{DBLP:journals/corr/VaswaniSPUJGKP17}, for the stock volatility prediction.

This study applies an attention-based LSTM to perform two news-driven stock prediction tasks, i.e. stock trend prediction and crisis early warning. With the established predictive model, predictions obtained from the DRNews embeddings and five baseline embedding models are compared and discussed. The primary objectives are (I) to evaluate the effectiveness of DRNews presentations against other news embedding paradigms in the context of news-driven stock market prediction, at the same time (II) to extract key information that drives stock market evolution from the news corpus. 

In remaining part of the paper is organized as follows. Section \ref{DRNews} describes the formation of the proposed DRNews model in terms of the construction of the news network and the implementation of news embeddings using the Subnode model. Section \ref{lstm} provides an overview of the attention-based LSTM model for stock predictions. In Section \ref{experiment}, the two stock market prediction experiments are presented. And Section \ref{conclusion} summarizes the conclusion. 

\section{The DRNews Model}\label{DRNews}
\begin{figure}[h!]
\centering
\includegraphics[scale=0.7]{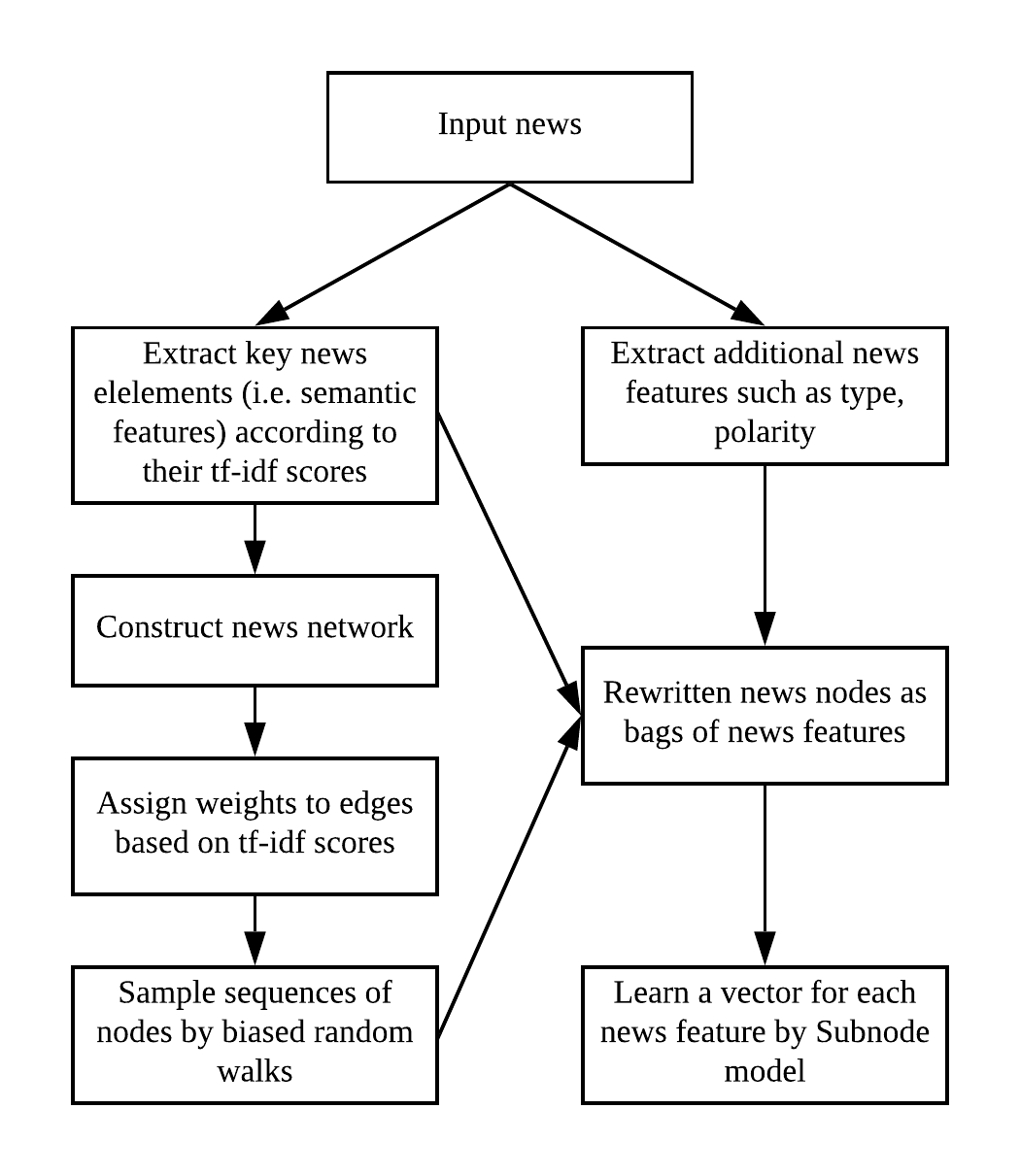}
\caption{Flow chart of the DRNews model}
\label{flow}
\end{figure}

In this section, the formation of the DRNews model is discussed. Figure \ref{flow} shows the overall flow chart of creating news embeddings with the DRNews model. With the text data of news articles as inputs, we first extract the key elements of news according to their term frequency-inverse document frequency (tf-idf) scores. The event elements are then collected as the news features together with some labeled features, such as publish date, words count, type and sentiment polarity, that are obtained from either directly the database or a trained classifier if there is a lack of ready-made labels. Based on these news elements and their tf-idfs, the news network is thus built whereas node sequences are sampled by the biased random walk. News nodes in these sequences are then represented as bags of extracted news attributes, a.k.a. features, and associated vectors are assigned to each features using the Subnode model.

In the rest of this section, a more explicit explanation is given in terms of the news network construction and the Subnode model for attributed network embedding. 
\subsection{Constructing the news network}
The DRNews model determines news embeddings based on an attributed news network, which is constructed by connecting the news node with its element nodes and news features as the news node attributes. With respect to the argument that a piece of news often describes one or more events that could be represented as a collection of event elements, the DRNews model extracts news elements as entities, actions and nouns by ranking the words/phrases by their tf-idf scores and select those with the $1\%$ highest tf-idfs. To comprise a news event, an entity is usually a person's name, an organization, or a location, whilst an action is often the trigger of an event at the same time contains key verbs that possibly reflect the type of the event. To enrich the news representation and address the problem that some verbs are often used as gerunds, we also consider nouns with high tf-idf scores as news elements. The tf-idf score is expressed as: 
\begin{equation}
    tfidf_{e,k}=\frac{n_{e,k}}{\sum_{w}n_{w,k}}\times\log\frac{\left|D\right|}{\left|{k:e \in d_{k}}\right|}
\end{equation}
where $n_{e,k}$ denotes the number of the news element $e$ in the news story $k$ and its denominator is the total number of words/phrases in the news document, $\left|D\right|$ is the total number of documents, the denominator represents the number of stories containing the element $e$. Overall, the tf-idf assigns a higher score to the element which appears more frequently in the news story than in others.

 Figure \ref{fig:network} gives a glance on the fundamental structure of the DRNews news network. Specifically, news nodes $k \in V_{k}$ are represented by their titles and the published time with no edges connecting any pair of element nodes $e \in V_{e}$ or news nodes. The weights of edges $(k,e)$ are assigned based on the importance of the event element to the news. The tf-idf is then used to measure the importance of a word/phrase inside the document and to determine the weight of the edge $(k,e)$:
\begin{equation}
    W_{k,e}=\frac{tfidf_{e,k}(title)}{Z_{1}}+\frac{tfidf_{e,k}(content)}{Z_{2}}
\end{equation}
The weight is simply the summation of the normalized tf-idfs of the event elements in the news title and text. Since there are fewer event elements in the title, the normalized weights tend to be higher than those in the content. This result is also true because headlines often contain the main information of an entire story.

\subsection{News network embedding with the Subnode Model}
\subsubsection{Network embedding by random walks}
To determine the news vectors, our network embedding approach shares a similar idea with the Node2vec model \cite{Grover:2016:NSF:2939672.2939754}. The goal is to learn a latent feature vector $F(v)$ for each node $v \in V$, that maximizes the probability of predicting the node $v's$ network neighborhood $N(v)$. Hence, the objective function is written as:
\begin{equation}
    max\sum_{v \in V}\log Pr(N(v)|F(v))
\end{equation}
Given the feature vector of the node $v$, it is assumed that the prediction probabilities of its neighborhoods are independent of one another, so
\begin{equation}
    logPr(N(v)|F(v)) = \sum_{u \in N(v)}logPr(u|F(v))
\label{eq:ass}
\end{equation}
where network neighborhoods $N(v)$ for each node are sampled by the biased random walk and concatenated into sequences. These sequences are processed by extending the skip-gram model \cite{DBLP:journals/corr/abs-1301-3781} from dealing with word sequences to node sequences. Specifically, negative sampling \cite{DBLP:journals/corr/MikolovSCCD13} is used to approximate the conditional probability:
\begin{equation}
\begin{split}
 logPr(u|F(v)) \approx logPr(1|F(v),u) + \\
\sum_{i=1}^m \mathbf{E}_{u^{'}_i \sim P(v)}[logPr(0|F(v),u^{'}_i)]
\end{split}
\label{eq:ng}
\end{equation}
where $u$ is sampled from the context of node $v$ in the sequence while $u^{'}$ ($u^{'} \neq u$) are sampled $m$ times from the noise distribution $P(v)$, which is typically the Unigram Distribution. Each sampling is treated as a simple binary classification to be solved by the corresponding logistic regression classifier.

Different from uniform random walks of DeepWalk \cite{Perozzi:2014:DOL:2623330.2623732}, Node2vec introduces two classic search strategies, namely the Breadth-first Search (BFS) and Depth-first Search (DFS). Suppose that the walk just transitioned from $t$ to $v$ and is now walking to its next step node $x$, in our network, there are two possible situations for the next step. One is revisiting $t$, {\em i.e.}, $d_{tx}=0$, whereas the other is keeping going deeper and further away from the node $t$, {\em i.e.}, $d_{tx}=2$. Node2vec defines a search bias $\alpha$ to control the search preference:
\begin{equation}
    \alpha_{vx}=\left\{
             \begin{array}{lr}
             \frac{1}{p},& if \ d_{tx}=0 \\
             \frac{1}{q},& if \ d_{tx}=2
             \end{array}
\right.
\label{eq:search}
\end{equation}
Next, the transition probability of edge $(v,x)$ is defined as 
the product of the weight of the edge and the bias, {\em i.e.}, $\frac{\alpha_{vx}*w_{vx}}{Z}$ where $Z$ is the normalization constant. If $p<q$, the walk is width-first associated with a greater probability to revisit the original node. Therefore, it is easier to sample nodes around the original node, that is, semantically similar news as they share many same news elements. If $p>q$, the walk is depth-first, and tends to visit nodes that have not been passed before. As such it is more likely to explore news with latent relations. During random walks, each node is used as a starting vertex for a fixed-length walk, producing a group of sequences. Based on the skip-gram model, node representations are finally trained on these sequences by optimizing the objective function using the Stochastic Gradient Descent (SGD).

\subsubsection{The Subnode model}
To address the major limitations that Node2vec can neither infer unseen nodes easily nor integrate node attributes into node representations, we propose a Subnode model to embed the DRNews network. 

Essentially, the Subnode model is inspired by the Subword model \cite{2016arXiv160704606B}, which represents each word using a bag of character $n-$grams. As an example, the word $where$ is represented as  $G=<wh,whe,her,ere,re>$. Based on the skip-gram architecture, each character $n-$gram can learn a vector representation and the word vector is the summation of its character $n-$grams.

By the biased random walk, sequences of news and event element nodes (such as $news_A \to element_A \to news_B \to element_B \to \cdots $) are obtained. Similar to Subword, the proposed Subnode rewrites each node as a bag of node attributes. With the focus on the news nodes, the node attributes are news features including semantic features such as event elements with high tf-idf scores, text structure features such as words count and paragraphs count and side information such as published date, news type and news emotion. Each news vertex $e \in V_e$ is thus represented as:
$$G=<entity_A,entity_B,action,\cdots,$$
$$words \ count,paragraph \ count,$$
$$month,day,week,type,sentiment,\cdots>$$
where words count, paragraph count and sentiment are ordinal data rather than real values. Similarly, the representation of a news node no longer directly corresponds to a distinct vector, but is instead the sum of the vectors of all its features. The objective function is now updated to:
\begin{equation}
    max\sum_{k \in V_k}\log Pr(N(k)|\sum_{g \in G_k}F(g))
    \label{eq:up}
\end{equation}

where $F(g)$ denotes the vector representation of a news feature $g$. It is worth-mentioning that nodes with only one edge are removed to reduce the sparsity of the network. The formulation of Eq. (\ref{eq:up}) is similar to Eq. (\ref{eq:ass}), (\ref{eq:ng}) and the vector representations of news features are trained by optimizing the updated objective function using SGD with negative sampling \cite{DBLP:journals/corr/MikolovSCCD13}. The pseudo-code of the Subnode algorithm is given in Algorithm 1.

In this way, the Subnode representation of unseen news nodes can be computed easily by extracting its news features, looking up for the corresponding vectors in the $feature \ to \ vector$ dictionary, and summing the vectors up. The algorithm code is open source and avaiblable at Github\footnote{https://github.com/yema2018/News2vec}.

\begin{algorithm}
\caption{Subnode}
\KwIn{$S$, the node sequence $S = [k^1,e^1,k^2,e^2,\cdots]$ generated by random walks}
\KwOut{$F$, the embedding matrix with respect to news feature $g$}
Rewrite the sequence as $S^{'} = [(g_1,g_2,\cdots), e^1, (g_3,g_4,\cdots), e^2, \cdots]$\;
Initialize $F$\;
\For{$s$ in $S'$, $s \neq e$}
{\For{$u$ in $Context(s)$}
{$NegativeSampling \ u'$ from $S'/\{s\}$\;
$J(F) = - logPr(u|\sum_{g \in G_s}F(g))$\;
$F := F - \eta * \frac{tial J}{tial F}$}}
Return $F$
\end{algorithm}

\subsection{Visualization of DRNews embeddings: dimension reduction plots}
\begin{figure*}[h!]
\centering
\includegraphics[scale=0.7]{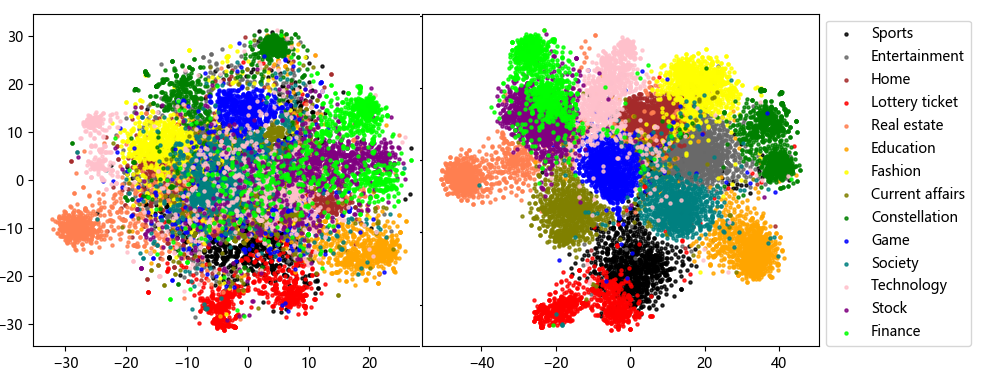}
\caption{Dimension reduction plots of test set news without and with the type feature}
\label{fig:thuc}
\end{figure*}

To give a more straightforward view on the DRNews embeddings, we train news vectors on the THUCTC\footnote{http://thuctc.thunlp.org} (THU Chinese Text Classification) news data set with one labeled feature, namely $type$, beside the semantic features. There are 14 types of news in the news corpus and we randomly take 6000/1000 pieces of news from each type to make up the training set/test set. Based on the news training set, a vector representation of 128 dimensions is learned for each news feature. Figure \ref{fig:thuc} shows the dimension reduction plots of news vectors in the test set using the t-Distributed Stochastic Neighbor Embedding (t-SNE). 
The right panel of Figure \ref{fig:thuc} shows the news vectors trained with the type feature, in which clustering patterns are distinctly evident, suggesting that DRNews allows the embedding of the type information into news vectors after incorporating the corresponding feature. Furthermore, it is observed that the distance between two clusters is positively correlated with the actual relevancy between the topics. In particular, neighboring relationships are detected between pairs of related topics, such as sports \& lottery ticket, current affairs \& society, stock \& finance, and fashion \& entertainment, whilst the distances between irrelevant clusters, such as constellation \& finance, game \& education, are larger. Nevertheless, the left panel of Figure \ref{fig:thuc} shows that news embeddings significantly lose classification information when vectors of type features are removed from the news vector. In a nutshell, the dimension reduction visualization of news vectors shows that our DRNews embedding model enables news vectors to represent multiple attributes simultaneously.

\section{The Deep Predictive Model}\label{lstm}
\begin{figure}[h!]
\centering
\includegraphics[scale=0.6]{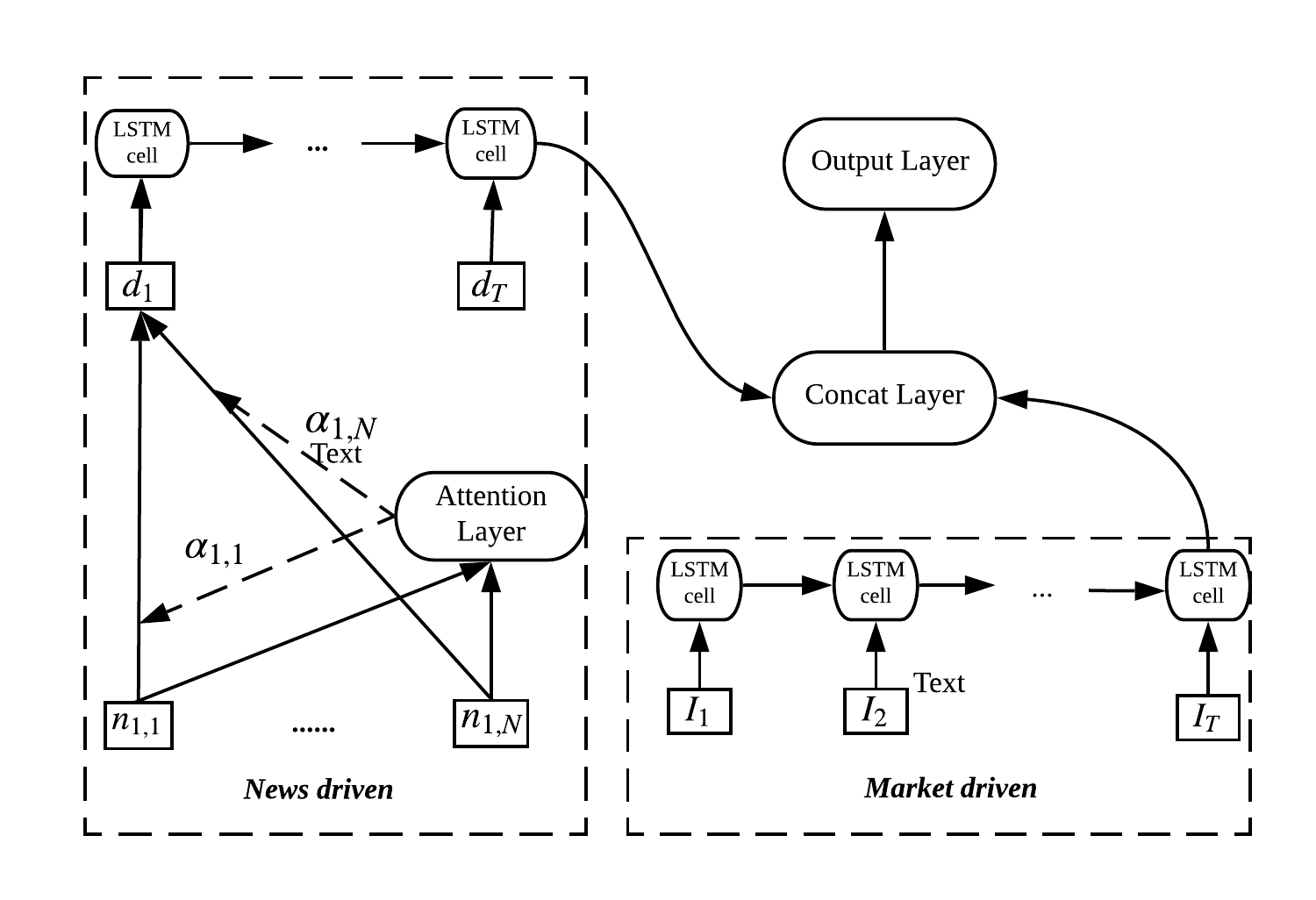}
\caption{The deep predictive model}
\label{fig:conc}
\end{figure}

This section discusses the deep predictive model that is used in the stock market prediction tasks by mining information from past news and market indicators. Since both news and market indicators can be viewed as daily time series, so a sequence model based on the long short-term memory (LSTM) network \cite{doi:10.1162/neco.1997.9.8.1735}) network is employed to encode the time series and to learn the effect of past news and market data on the future market. As different news stories tend to have different relevance to the market, an attention mechanism \cite{N16-1174} is adopted to assign a weight to each news automatically by gradient descent. The prediction structure is shown in Figure \ref{fig:conc} as a bi-input neural network for two types of data, namely news texts and the numeral data of historical returns. In particular, the news data is encoded by the attention-based LSTM, while the market data is encoded by a sole LSTM model. Their outputs at the final time step are concatenated and fed into the output layer.

\textbf{Attention model}: To make news-driven stock predictions, each news vector $n_{t,i}$ is first fed into a single-layer network, whose outputs $u_{t,i}$ are normalized by a softmax activation function and weighted ($\alpha_{t,i}$) by the corresponding news. Specifically,
\begin{equation}
    u_{t,i} = tanh(W_nn_{t,i}+b_n) 
\end{equation}
\begin{equation}
    \alpha_{t,i} = \frac{exp(u_{t,i}^Tu_w)}{\sum_j{exp(u_{t,j}^Tu_w)}} 
\end{equation}
\begin{equation}
    d_t = \sum_i{\alpha_{t,i}n_{t,i}}
\end{equation}
where $W_n$ are initialized at the beginning and trained by back propagation. After determining weights of all the news, the daily news vector $d_t$ is computed as the weighted sum of all the news vectors in that day, integrating the daily news information. 

\textbf{LSTM model}: The LSTM model is a variant of the recurrent neural network that uses updating $\Gamma_u$ and forgetting $\Gamma_f$ factors to update the memory cell $c^{<t>}$. Specifically,
\begin{equation}
    \widetilde{c}^{<t>} = tanh(W_c[a^{<t-1>},x^{<t>}] + b_c)
\end{equation}
\begin{equation}
    \Gamma_u = sigmoid(W_u[a^{<t-1>},x^{<t>}]+b_u)
\end{equation}
\begin{equation}
    \Gamma_f = sigmoid(W_f[a^{<t-1>},x^{<t>}]+b_f)
\end{equation}
\begin{equation}
    \Gamma_o = sigmoid(W_o[a^{<t-1>},x^{<t>}]+b_o)
\end{equation}
\begin{equation}
    c^{<t>} = \Gamma_u\times\widetilde{c}^{<t>}+\Gamma_f\times c^{<t-1>}
\end{equation}
\begin{equation}
    a^{<t>} = \Gamma_o\times tanh(c^{<t>})
\end{equation}
where $ a^{<t-1>}$ is both the output at the time step $t-1$ and one of the inputs of the next LSTM cell. Another input of the LSTM cell is possibly $x^{<t>}$ which is an integration $d_t$ of news set $\{n_{t,1},\cdots,n_{t,N}\}$ for the news-driven LSTM, or a vector $I_t$ of historical returns for the market-driven LSTM. In addition to the output of $a^{<:>}$, the memory factor $c^{<:>}$ is passed through the LSTM cells along with the output being constantly updated to ensure that the most important information is retained.

\textbf{Training}: Encoded News and market data are integrated by a concatenation layer and fed into an output layer with $J$ neurons. Note the number $J$ depends on how many classifications a task needs to discriminate. We train the $J$ classifier by optimizing the cross entropy loss with the Adam optimization algorithm,
\begin{equation}
    L(\hat{y},y) = -\sum_{j=1}^J{y_jlog\hat{y}_j} + \lambda\left\|Q\right\|_2^2
\end{equation}
where the $Q$ is the set of all trainable parameters and $\lambda$ is the $L_2$ regularization weight. Through the concatenation, parameters from both news-driven and market-driven model are updated simultaneously by the gradient descent.

\textbf{Performance measures}: To investigate the forecasting power of the attention-based LSTM model, two performance measures, i.e. the Accuracy (Acc) and the Matthews Correlation Coefficient (MCC), are hired. In particular, the accuracy of a binary predictive model takes value in the interval between $0$ and $1$ and is computed as:
\begin{equation}
\text{Acc} = \frac{\text{TP + TN}}{\text{TP + TN + FP + FN}},
\end{equation}
where TP = True positive, FP = False positive, TN = True negative, FN = False negative. As a result, models with higher values of accuracy are preferred as a higher level of predicting power is indicated. On the other hand, the MCC measure is adopted as the the quality measure of the predictive model, which returns the correlation coefficient between the observed and predicted classifications: 
\begin{equation}
\text{MCC} = \frac{\text{TP} \times \text{TN} - \text{FP} \times \text{FN}}{\sqrt{\text{(TP + FP)(TP + FN)(TN + FP)(TN + FN)}}}.
\end{equation}

The MCC measure is considered as one of the most reliable criteria in the selection of classification models for its strength in dealing with imbalanced data. The value of the MCC ranges from $-1$ to $1$, where a perfect classification model is indicated by $1$, and a completely wrong model is indicated by $-1$. 

\section{Experiments}\label{experiment}
\subsection{Stock Movement Prediction}
\subsubsection{Experimental settings}
We use financial news (2009/10/19 to 2016/10/31) from Sohu\footnote{https://www.jianshu.com/p/370d3e67a18f} to predict the daily movement of Shanghai Securities Composite (SSE) index. 

To obtain the news embeddings, vectors of the news features are first obtained according to the news network from 2009/10/19 to 2015/12/31 whereas news vectors are then computed by summing up its associated feature vectors. In particular, the news network is trained by setting the length of the biased random walk as $100$, the context size as $10$, the return hyper-parameter ($p$ in Eq. (\ref{eq:search})) and the input hyper-parameter ($q$ in Eq. (\ref{eq:search})) both as $1$. Beside the semantic features, five additional features, that attach the basic information of time (three features: $month$, $day$, $week day$), news sentiment polarity (one feature: $sentiment$)\footnote{In this study, the sentiment polarity, is determined by the number of positive words over the number of negative words.} and words count (one feature: $words\ count$) are incorporated for each piece of news considered. 

Next, we use the previous 20 days ($T=20$) of news and market returns to predict the next day's movement of the SSE stock index. The output layer is a ternary classifier to indicate whether the index moves UP (daily return $>$ $0.33\%$), DOWN (daily return $<$ $-0.29\%$) or PRESERVES ($-0.29\%$ $<$ daily return $<$ $0.33\%$). We employ a rolling window with $size=20$ and $strider=1$ to augment the sample.

In this study, news and market data before the year 2016 is used as the training set, whilst the rest is used as the test-validation set. 

\subsubsection{Stock movement prediction results}
 As the baseline models, we include the following news embedding approaches that are used for stock movement predictions. Note that to ensure the validity of comparison, the performance measures of all models are computed from the same input data of daily news text. The other two event embedding models, namely Event2vec \cite{Setty:2018:ENE:3209978.3210136} and knowledge-enhanced event embedding \cite{Ding2016KnowledgeDrivenEE}, are not considered as the former could only be used in an organized event data set and the latter requires additional knowledge graphs.\\
\textbf{- Average BOW}: News is represented as the average of word vectors in the news titles and bodies \cite{Hu:2018:LCW:3159652.3159690,Das:2017:EEF:3132847.3133131}. Word embedding is obtained by training the skip-gram Word2vec model \cite{DBLP:journals/corr/abs-1301-3781}. \\
\textbf{- Doc2vec} \cite{DBLP:journals/corr/LeM14}: News text is represented as dense vectors by the Paragraph Vector model \cite{7550882}.\\
\textbf{- Event embedding} \cite{Ding:2015:DLE:2832415.2832572}: Event tuples are extracted from headlines and represented as vector representations by scoring the correct tuples higher than the corrupted ones.\\
\textbf{- Deepclue} \cite{article}: Different from the two-stage prediction of DRNews that returns the news representation and the prediction separately, Deepclue adopts an end-to-end architecture to predict next day's stock price using words in headlines from the day before. To use Deepclue as a baseline of creating news embeddings, we modify the model to a two-stage task. Specifically, we first use Deepclue to train a news encoder, instead of predicting stock prices. After the news representations are obtained from the trained encoder, we then feed the news vectors into our attention-based LSTM model to make stock market predictions. In this study, we only include two out of three news encoders proposed by Deepclue, namely the CNN and LSTM, as the third encoder based on the bi-gram produces no trainable parameters. \\
In addition, we apply a sentence-level encoder BERT \cite{DBLP:journals/corr/abs-1810-04805} to encode news headlines as another baseline model:\\
\textbf{- BERT}: Average pooling is used to get a fixed representation of a headline based on the Chinese pre-trained BERT model\footnote{https://github.com/google-research/bert\#pre-trained-models}.\\
The proposed DRNews model is implemented with and without the five additional, {\em i.e.} $month$, $week$, $sentiment$, $words\ count$ and $day$, features:\\
\textbf{- DRNews-}: News vectors are the summation of solely the semantic features ({\em i.e.}, event elements).\\
\textbf{- DRNews+}: News vectors are the summation of all news features including the semantic and the five additional features.
\begin{table}[h] 
\centering  
    \begin{tabular}{lcc}
        \hline \hline &Acc&MCC\\
        \hline
        Average \ BOW ($d=128$)&$41.91\%$&$0.1137$\\
        Doc2vec ($d=128$)&$43.90\%$&$0.1484$\\
        Event \ embedding ($d=128$)&$43.76\%$&$0.1439$\\
        Deepclue-LSTM ($d=128$)&$41.04\%$&$0.1034$\\
        Deepclue-CNN ($d=128$)&$41.43\%$&$0.1066$\\
        BERT ($d=764$)&$45.83\%$&$0.1442$\\
        DRNews- ($d=128$)&$44.68\%$&$0.1664$\\
        DRNews+ ($d=128$)&$\mathbf{46.24\%}$&$\mathbf{0.1936}$\\
        \hline
        Market Return&$41.04\%$&$0.1050$\\
        \hline
        DRNews+ \& Return&$44.22\%$&$0.1637$\\
        \hline
        \hline
    \end{tabular}
\caption{Acc and MCC results of stock movement predictions in the test set} 
\label{res}
\end{table}

In Table \ref{res}, two performance measures, i.e. the Acc and the MCC, are given to describe the predicting power of DRNews in comparison to other news embedding approaches. Specifically, the first panel lists the Acc and MCC of predictions based on each individual embedding model, the second panel gives the prediction result solely using the historical data of return, whilst the last panel shows the prediction with the joint information of the historical return and the DRNews embedding. Major implications of Table \ref{res} are threefold. 

First, it is shown that stock movement predictions solely based on the embedded information of news (Panel 1) outperform the predictions taking into account the historical data of market return (Panel 2 and 3). This finding suggests that daily news embeddings contain sufficient information for short-term stock movement prediction, at the same time indicates that the SSE stock market complies the weak-form of efficiency as the historical price data makes no contribution to the prediction of its future movements. 

Second, the DRNews model outperforms the baseline models with the most prominent Acc and MCC results in a lower dimension of $d=128$ \footnote{In the experiment of stock movement prediction, the three best-performing models are DRNews+ (higest Acc and MCC), BERT (Second highest Acc) and DRNews- (Second highest MCC). The dimension of BERT is $764$.}. As for news-driven stock price prediction, it is more reasonable to take into account potential connections between news, such as sharing similar background story or leading to the same event, and reflect the connection by embedding similar values in certain dimensions of the news vector. Through the biased random walk in the news network, DRNews embeddings capture inter-textual knowledge, {\em i.e.} latent connections between news, which enriches the news vectors with vital information that contributes to the high predictive power of the model. On the other hand, Doc2vec, Event embedding and BERT produce stock movement predictions of similar qualities in terms of MCC, as the news vectors generated by the three approaches uniformly limit to the textual information within news articles. 

Last, the DRNews algorithm is competent to enrich the news vector representation with additional information brought by supplementary features, as the Acc and MCC results of DRNews+ improve by $1.56\%$ and $2.72\%$ in comparison to DRNews- after incorporating the five labeled features ($month$, $week$, $day$ $sentiment$, and $words\ count$). It is worth-mentioning that in this study, the New2vec model is performed in an unsupervised manner as all features are extracted from common news text without any artificial label or trained classifier. Therefore, we believe that there is a large chance to further improve the result once higher quality labels, such as news sentiment from trained classifier or subcategories of financial news, are incorporated. Despite that the performance of Deepclue is hardly satisfactory, we believe that this supervised news embedding model has great potential once a stronger news encoder and supervised tasks are placed to enhance the capacity of feature extraction and the universality of embedding.

\subsection{Early Warning of Stock Market Crises}
\subsubsection{Experimental settings}
In this section, we use an attention-based LSTM network to construct an early warning model for stock market crises. To classify the crisis/non-crisis dates of the SSE index, the switching ARCH (SWARCH) model \cite{hamswar} is hired through the modeling of high/low volatility regimes that reflect respectively the tranquility and turmoil \cite{wang2019integrated}. To determine whether a crisis occurs, a SWARCH(2,1) process is written as: 
\begin{align}
y_{t} &= u + \theta_{1}y_{t-1}+\epsilon_{t},\quad \epsilon_{t}|\mathcal{I}_{t-1}\sim N(0, h_{t});\label{swarch1}\\
\frac{h_{t}^{2}}{\gamma_{s_{t}}} &= \alpha_{0}+\alpha_{1}\frac{\epsilon_{t-1}^{2}}{\gamma_{s_{t-1}}}, s_{t}=\{1,2\}.\label{swarch2}
\end{align}

Eq.(\ref{swarch1}) describes an AR(1) process with a normal error term $\epsilon_t$ of variance $h_{t}$. The regime switching structure of the residual variance $h_{t}$ is given by Eq.(\ref{swarch2}) where the $\alpha's$ are non-negative, the $\gamma's$ are scaling parameters that capture the change in each regime, $s_{t}$ is the state variable that $s_{t}=1$ indicates the low volatility state of non-crisis, and $s_{t}=2$ indicates the high volatility state of crisis. 

The classification of high/low volatility regimes can be implemented on the basis of the so-called filtering probability, which is a byproduct of the maximum likelihood estimation. The filtering probability, written as:
\begin{align}
P(s_{t}=i|Y_{t};\boldsymbol{\theta})
\end{align}
where $\boldsymbol{\theta}$ is the vector of all model parameters to be estimated, could be interpreted as the conditional probability based on the current information of time $t$. The crisis classifier is defined as the following binary function with a probability threshold $0.5$ \cite{hamsus,hamlin}:
\begin{equation}
\text{Crisis} = 
\begin{cases}
1, & \text{SWARCH filtering probability}  \geq 0.5 \\
0, & \text{otherwise.}
\end{cases} 
\end{equation}	

\begin{table*}[!h]
	\caption{Summary of test-set forecasting. }	
	\setlength{\tabcolsep}{15pt}
	\centering
	\small
    \begin{tabular}{cccc}
            \hline
            &True crises&Market return &DRNews+ \& return\\
            & &predictions& predictions\\
            \hline
            &&&\\
            Total onsets&2&2&2\\
            Forewarned onsets&&1&2\\
            \% of forewarned onsets&&50&100\\
            Avg. days-ahead onsets &&1.0&2.5\\
      	  	\hline
	\end{tabular}\label{tab:correct}	
\end{table*}
By inputting news embeddings and historical price data, we aim to investigate the performance of DRNews in news-driven crisis predictions in comparison with other embedding models. Similar to the stock trend prediction, we use the news vectors and market return of the first 20 days to predict whether a stock market turbulence will occur in the next day. The difference is that the crash warning system is basically a binary classifier. Besides, a rolling window with $size=20$ and $strider=1$ is employed to augment the sample. 
Finally, the time interval for training is between 2009/10/19 and 2013/4/25 and the interval for testing is from 2013/4/26 to 2016/10/31.

\subsubsection{Crises early warning results}
\begin{table}[h] 
\centering  
    \begin{tabular}{lcc}
        \hline \hline &Acc&MCC\\
        \hline
        Average \ BOW ($d=128$)&$66.23\%$&$0.4517$\\
        Doc2vec ($d=128$)&$68.74\%$&$0.4992$\\
        Event \ embedding ($d=128$)&$69.82\%$&$0.4985$\\
        Deepclue-LSTM ($d=128$)&$49.40\%$&$0.0828$\\
        Deepclue-CNN ($d=128$)&$70.06\%$&$0.4443$\\
        BERT ($d=764$)&$74.61\%$&$0.4635$\\
        DRNews- ($d=128$)&$69.58\%$&$0.512$\\
        DRNews+ ($d=128$)&$\mathbf{77.13\%}$&$\mathbf{0.5681}$\\
        \hline
        Market Return&$94.25\%$&$0.8776$\\
        \hline
        DRNews+ \& Return&$\mathbf{94.61\%}$&$\mathbf{0.8861}$ \\
        \hline
        \hline
    \end{tabular}
\caption{Acc and MCC results of stock crisis predictions in the test set} 
\label{res2}
\end{table}

According to the Acc and MCC results of the six embedding approaches shown in Panel 1, DRNews+ remains as the best-performing embedding model in terms of both Acc and MCC, whilst BERT produces the second highest Acc and DRNews- produces the second highest MCC. As a concluding remark, the DRNews model demonstrates robust and strong performances in the completion of the two financial prediction tasks, i.e. stock movement prediction and stock crises warning, that requires the representation of latent linkages between news events. 

On the other hand, Panel 2 and 3 show that the results of crises warning is significant improved after historical return data is employed. Despite that the market-driven model already offers a fairly good prediction with $94.25\%$ Acc and $87.76\%$ MCC, the joint prediction of DRNews embedding and historical return generates the highest Acc and MCC of $94.64\%$ and $88.61\%$, respectively.

\begin{figure}[h!]
\centering
\includegraphics[scale=0.5]{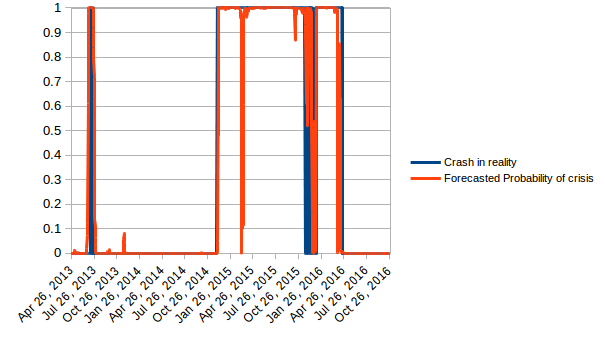}
\caption{The SWARCH filter probability of crash warning from historical return}
\label{fig:sig}
\end{figure}

\begin{figure}[h!]
\centering
\includegraphics[scale=0.5]{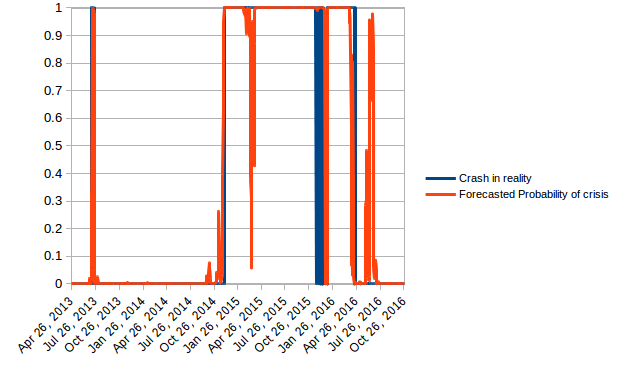}
\caption{The SWARCH filter probability of crash warning from historical return and DRNews+}
\label{fig:bi}
\end{figure}

\begin{CJK}{UTF8}{gbsn}
\begin{table*}[h] 
\centering  
    \begin{tabular}{ll}
        \hline \hline 
        \multicolumn{2}{c}{Important nouns}\\
        \hline
        Stock Trend (Ranking) & Crisis Warning\\
        \hline
        A股/A-share (1) & 银行/Bank (3)\\
        创业板/GEM (Growth Enterprise Market) (4) & 利率/Interest rate (5)\\
        基金/Fund (5) & CPI/CPI (9) \\
        ST/Special Treatment stock (8) & 存款/Deposit (11) \\
        上市公司/Listed company (10) & 房地产/Real estate (14) \\
        期货/Futures (11) & 不良贷款/Non-performing loans (24)\\
        券商/Brokerage (18) & 准备金率/Reserve ratio (35)\\
        沪指/Shanghai stock index (20) & 公积金/Provident fund (40)
        \\
        CPI/CPI (25) & 信托/Trust (41)\\
        限售股/Restricted shares (33) & 上市公司/Listed companies (46)\\
        IPO/IPO (36) &  光伏/Photovoltaic (53)\\
        融资融券/Margin financing (45) & 货币政策/Monetary policy (55)\\
        ETF/Exchange Traded Funds（47）& 流动性/Liquidity (65)\\
        市值/Market capitalization (51) & PMI/Purchase Managers' Index (73)\\
        新三板/New OTC(Over the Counter) Market (64) & 中小企业/Middle and small-sized enterprises (80)\\
      
        \hline
        \hline
    \end{tabular}
    \caption{Nouns with High Attention (Partial Selection)}
    \label{tab:noun}
\end{table*}

\begin{table*}[h] 
\centering  
    \begin{tabular}{ll}
        \hline \hline 
        \multicolumn{2}{c}{Top companies and institutes}\\
        \hline
        Stock Trend (Ranking) & Crisis Warning\\
        \hline
        证监会/China Securities Regulatory Commission (13) & 央行/People's Bank of China(8)\\
        光大证券/Everbright Securities (89) & 万科/Vanke (43)\\
        深交所/Shenzhen Stock Exchange (104) & 银监会/China Banking Regulatory Commission (49) \\
        上交所/Shanghai Stock Exchange (114) &保利地产/Poly Real Estate (101)  \\
        
        广发证券/GF Securities (116) & 浦发银行/Shanghai Pudong Development Bank (106) \\
        中信证券/CITIC Securities (126) &农行/Agricultural Bank of China (112)\\
        中金所/China Financial Futures Exchange (128) & 民生银行/China Minsheng Bank (143)\\
        华泰证券/Huatai Securities (154) &  中行/Bank of China (155)\\
        兴业证券/Industrial Securities (159) & 建行/China Construction Bank (162)\\
        招商证券/China Merchants Securities (160) & 中信银行/China CITIC Bank (170)\\
        纳斯达克/NASDAQ (169) &  国美/Gome (181)\\
        西南证券/Southwest Securities (170) & 平安银行/Ping An Bank (182)\\
        金融时报/Finance Times (176) & -\\
        海通证券/Haitong Securities (183) & - \\
        纽约商品交易所/New York Mercantile Exchange (194)& - \\
        \hline
        \hline
    \end{tabular}
    \caption{Company and Institute Names with High Attention}
    \label{tab:com}
\end{table*}

Figure \ref{fig:sig} and \ref{fig:bi} show the predicted SWARCH filter probability of crises based on the historical return with and without DRNews embeddings, respectively. Despite that the market-driven crises warning based on solely the historical return achieves impressive results in terms of Acc and MCC measures, it creates a delay of warning (see Oct 26, 2014 to Jan 26, 2015, red line in Figure \ref{fig:sig}) in comparison to the actual occurrence of the crisis (blue line in Figure \ref{fig:sig}). On the other hand, Figure \ref{fig:bi} suggests that the joint prediction of DRNews embedding and historical return generates true 'early' warning signals with predicted probability of crisis gradually raises at the end of 2014 before the actual crash. It appears that DRNews embeddings contain key information that serves as the driving force of crisis alarming during the pre-turmoil episode. In the next section, the key information inherited in the news will be discussed explicitly according to the attention-based LSTM model based on the DRNews embeddings. 

Table \ref{tab:correct} confirms the findings in Figure \ref{fig:sig} and \ref{fig:bi}. During the test-set period between April 2013 to October 2016, two stock crisis onsets are detected by the SWARCH model. The joint predictions based on the historical return and DRNews embeddings forewarn both onsets with an average of $2.5$-days forewarned period, whereas predictions with solely the historical return demonstrate a delay in warning the second onset that take place at the end of 2014.

\subsection{News Attention}

\begin{table}[h] 
\centering  
    \begin{tabular}{ll}
        \hline \hline 
        \multicolumn{2}{c}{Top actions (verb \& gerund)}\\
        \hline
        Stock Trend (Ranking) & Crisis Warning\\
        \hline
        增长/Growth (3) & 增长/Growth (1)\\
        披露/Disclosure (16) & 贷款/Loan (2)\\
        交易/Transaction (17) & 下降/Decline (4) \\
        上涨/Raising (21) & 信贷/Credit (9) \\
        解禁/Unban (24) & 融资/Financing (16) \\
        退市/Delisting (26) & 上涨/Rising (23)\\
        下跌/Falling (27) & 回落/Fallback (32)\\
        复牌/Resumption (28) & 投资/Investment (33)
        \\
        停牌/Suspension (32) & 调控/Regulate and control (37)\\
        投资/Investment (34) & 降息/Interest rate cuts (50)\\
        上市/Listing (41) &  投放Launch (61)\\
        公开发行/Public offering (68) & 降准/Reserve requirement ratio cuts (68)\\
        反弹/Rebound (80) & 出口/Export (71)\\
        亏损/Loss (82) & 成交/Close the deal (88) \\
        重组/Restructuring (88) & 加息/Interest rate increase (97)\\
        \hline
        \hline
    \end{tabular}
\caption{Actions with High Attention (Partial Selection)}
\label{tab:act}
\end{table}

The DRNews embeddings offer a mechanism that allows quantitative investigation of news-driven market predictions via the attention model. In this section, we select the daily news that obtains the highest attention weights in the stock movement prediction and crises warning models, respectively. The top 200 elements of the selected news are then extracted based on their tf-idf scores and listed in Table \ref{tab:act} - \ref{tab:com} as actions, nouns, and company\&institute names. By comparing the key elements in the two financial tasks, i.e. stock movement and crisis predictions, we aim to gain insights into the dynamics of the stock market as well as the information transmission channel between media and the financial sector. 

With respect to Table \ref{tab:act} - \ref{tab:com}, it is found that the predictions of short-term stock movement and stock market turbulence exhibit different concentrations, with the former receiving more impacts directly related to the secondary market whereas the latter exposed under wider influences from the economical/regulatory developments as well as the banking system. In particular,

Table \ref{tab:act} shows the key actions in news that are highly correlated with the stock movement (left panel) and the stock market turbulence (right panel). In the case of stock movements, \textbf{\em Disclosure} appears to be rather frequently quoted in high-attention news, indicating that the stock dynamics is information-driven with a great level of sensitivity. In addition, operations on listing companies, such as \textbf{\em Delisting \& Listing} and \textbf{\em Resumption \& Suspension}, \textbf{\em Unban} and \textbf{\em Public offering}, are closely related to the stock price oscillation. By contrast, stock market crashes are greatly connected with news that has to do with the interest rate and the banking sector. Major keywords that drives crisis predictions include \textbf{\em Loan}, \textbf{\em Credit}, \textbf{\em Financing}, as well as phrases related to the regulatory changes in the interest rate and the reserve requirement, such as \textbf{\em Interest rate cuts}, \textbf{\em Reserve requirement ratio cuts} and \textbf{\em Interest rate increase}.

Clear differences between the important nouns in the stock movement and crisis predictions are observed in Table \ref{tab:noun}. Basically, the stock movement shows great responses to news directly related to the stock market with key mentions including the \textbf{\em A-share, Growth Enterprise Market, Shanghai stock index, Futures, ETF, Restricted shares, New OTC Market}, etc. On the other hand, the occurrence of the stock market turmoil is mainly reflected by the macroeconomic indicators and policies, such as the \textbf{\em Interest rate, CPI, Reserve ratio, Provident fund, Liquidity, PMI} and \textbf{\em Monetary policy}.

Table \ref{tab:com} of important company and institute names supports the claims made from Table \ref{tab:act} and \ref{tab:noun}. Specifically, the stock movement is found to be heavily affected by three types of institutions, namely securities companies, stock exchanges and securities regulators. On the contrary, stock market crashes are shown to be largely influenced by news related to the banking sector. Among the monetary-oriented institutes, two government agencies, i.e. \textbf{\em the People's Bank of China} (the Central Bank) and \textbf{\em the China Banking Regulatory Commission}, are involved demonstrating their leading effects on the systematic risk of the stock market, followed by seven commercial Banks including \textbf{\em Shanghai Pudong Development Bank, the Agricultural Bank of China, China Minsheng Bank}, etc. Additionally, two real estate enterprises (\textbf{\em i.e. Vanke \& Ploy Real Estate}) are frequently mentioned in news with high attention weights.

\section{Conclusions}\label{conclusion}

DRNews, a news embedding model with distributed representations is developed and applied for stock market predictions in this study. The DRNews model enhances the established news embedding methods by representing news articles in an attributed network that integrates the semantic information with the inter-textual knowledge. The information contained in the network covers not only the natural co-occurrence relationship between news and its internal event elements, but also latent connections between cross-documental news events. To determine the news embeddings in the framework of a network, a Subnode algorithm is then proposed to enrich the news vectors with various features at the same time be able to produce fast inferential embeddings of latest news outside the network. Note that despite the focus of Subnode model in this paper is its application in the news network, this algorithm could be widely applied to other networks inheriting its merit of providing an effective solution to embed unseen nodes and integrate all node attributes in the embedding.

Using the attention-based LSTM network, the DRNews embeddings are applied in two stock market prediction tasks, namely the short-term price movement prediction and stock crises early warning. By comparing with five baseline models, it is shown that DRNews achieves state-of-art performance in all aspects, mainly due to its comprehensive description of various news features and potential linkages among news events on the vector representations. In particular, DRNews produces a test-set accuracy of $46.24\%$ in the ternary prediction task of stock movement. As for the stock crisis early warning, a test-set accuracy of $94.61\%$ is obtained with an average of $2.5$-days forewarned period. Further, our experiments in the stock market show that daily news plays substantial roles on affecting both the short-term stock price oscillation and the systematic evolution of the stock market. In particular, the short-term stock movement is more easily influenced by secondary market-related information, whilst the stock crises are mainly merged from matters related to the banking sector and the economic system.

\section*{acknowledgements}
We acknowledge the support by XJTLU Key Program Special Fund-Applied Technology Research Programmme (No. KSF-A-14).
\end{CJK}

\newpage

\bibliographystyle{spmpsci}  
\bibliography{main}

\begin{thebibliography}{10}
\providecommand{\url}[1]{{#1}}
\providecommand{\urlprefix}{URL }
\expandafter\ifx\csname urlstyle\endcsname\relax
  \providecommand{\doi}[1]{DOI~\discretionary{}{}{}#1}\else
  \providecommand{\doi}{DOI~\discretionary{}{}{}\begingroup
  \urlstyle{rm}\Url}\fi

\bibitem{7550882}
{Akita}, R., {Yoshihara}, A., {Matsubara}, T., {Uehara}, K.: Deep learning for
  stock prediction using numerical and textual information.
\newblock In: 2016 IEEE/ACIS 15th International Conference on Computer and
  Information Science (ICIS), pp. 1--6 (2016).
\newblock \doi{10.1109/ICIS.2016.7550882}

\bibitem{2016arXiv160704606B}
{Bojanowski}, P., {Grave}, E., {Joulin}, A., {Mikolov}, T.: {Enriching Word
  Vectors with Subword Information}.
\newblock arXiv e-prints  (2016)

\bibitem{Bordes:2013:TEM:2999792.2999923}
Bordes, A., Usunier, N., Garcia-Dur\'{a}n, A., Weston, J., Yakhnenko, O.:
  Translating embeddings for modeling multi-relational data.
\newblock In: Proceedings of the 26th International Conference on Neural
  Information Processing Systems - Volume 2, NIPS'13, pp. 2787--2795. Curran
  Associates Inc., USA (2013).
\newblock \urlprefix\url{http://dl.acm.org/citation.cfm?id=2999792.2999923}

\bibitem{DBLP:journals/corr/abs-1803-11175}
Cer, D., Yang, Y., Kong, S., Hua, N., Limtiaco, N., John, R.S., Constant, N.,
  Guajardo{-}Cespedes, M., Yuan, S., Tar, C., Sung, Y., Strope, B., Kurzweil,
  R.: Universal sentence encoder.
\newblock CoRR \textbf{abs/1803.11175} (2018).
\newblock \urlprefix\url{http://arxiv.org/abs/1803.11175}

\bibitem{chatzis2018forecasting}
Chatzis, S.P., Siakoulis, V., Petropoulos, A., Stavroulakis, E.,
  Vlachogiannakis, N.: Forecasting stock market crisis events using deep and
  statistical machine learning techniques.
\newblock Expert systems with applications \textbf{112}, 353--371 (2018)

\bibitem{10.1007/978-3-642-24704-0_4}
Chen, H.: Predicting market movements: From breaking news to emerging social
  media.
\newblock In: A.~Datta, S.~Shulman, B.~Zheng, S.D. Lin, A.~Sun, E.P. Lim (eds.)
  Social Informatics, pp. 5--5. Springer Berlin Heidelberg, Berlin, Heidelberg
  (2011)

\bibitem{DBLP:journals/corr/ConneauKSBB17}
Conneau, A., Kiela, D., Schwenk, H., Barrault, L., Bordes, A.: Supervised
  learning of universal sentence representations from natural language
  inference data.
\newblock CoRR \textbf{abs/1705.02364} (2017).
\newblock \urlprefix\url{http://arxiv.org/abs/1705.02364}

\bibitem{DBLP:journals/corr/abs-1805-01070}
Conneau, A., Kruszewski, G., Lample, G., Barrault, L., Baroni, M.: What you can
  cram into a single vector: Probing sentence embeddings for linguistic
  properties.
\newblock CoRR \textbf{abs/1805.01070} (2018).
\newblock \urlprefix\url{http://arxiv.org/abs/1805.01070}

\bibitem{Das:2017:EEF:3132847.3133131}
Das, S., Mishra, A., Berberich, K., Setty, V.: Estimating event focus time
  using neural word embeddings.
\newblock In: Proceedings of the 2017 ACM on Conference on Information and
  Knowledge Management, CIKM '17, pp. 2039--2042. ACM, New York, NY, USA
  (2017).
\newblock \doi{10.1145/3132847.3133131}.
\newblock \urlprefix\url{http://doi.acm.org/10.1145/3132847.3133131}

\bibitem{DBLP:journals/corr/abs-1810-04805}
Devlin, J., Chang, M., Lee, K., Toutanova, K.: {BERT:} pre-training of deep
  bidirectional transformers for language understanding.
\newblock CoRR \textbf{abs/1810.04805} (2018).
\newblock \urlprefix\url{http://arxiv.org/abs/1810.04805}

\bibitem{di2016artificial}
Di~Persio, L., Honchar, O.: Artificial neural networks architectures for stock
  price prediction: Comparisons and applications.
\newblock International journal of circuits, systems and signal processing
  \textbf{10}, 403--413 (2016)

\bibitem{Ding:2015:DLE:2832415.2832572}
Ding, X., Zhang, Y., Liu, T., Duan, J.: Deep learning for event-driven stock
  prediction.
\newblock In: Proceedings of the 24th International Conference on Artificial
  Intelligence, IJCAI'15, pp. 2327--2333. AAAI Press (2015).
\newblock \urlprefix\url{http://dl.acm.org/citation.cfm?id=2832415.2832572}

\bibitem{Ding2016KnowledgeDrivenEE}
Ding, X., Zhang, Y., Liu, T., Duan, J.: Knowledge-driven event embedding for
  stock prediction.
\newblock In: COLING (2016)

\bibitem{fischer2018deep}
Fischer, T., Krauss, C.: Deep learning with long short-term memory networks for
  financial market predictions.
\newblock European Journal of Operational Research \textbf{270}(2), 654--669
  (2018)

\bibitem{Grover:2016:NSF:2939672.2939754}
Grover, A., Leskovec, J.: Node2vec: Scalable feature learning for networks.
\newblock In: Proceedings of the 22Nd ACM SIGKDD International Conference on
  Knowledge Discovery and Data Mining, KDD '16, pp. 855--864. ACM, New York,
  NY, USA (2016).
\newblock \doi{10.1145/2939672.2939754}.
\newblock \urlprefix\url{http://doi.acm.org/10.1145/2939672.2939754}

\bibitem{hamlin}
Hamilton, J., Gang, L.: Stock {M}arket {V}olatility and the {B}usiness {C}ycle.
\newblock Journal of Applied Econometrics \textbf{11}(5), 573--93 (1996).
\newblock
  \urlprefix\url{https://EconPapers.repec.org/RePEc:jae:japmet:v:11:y:1996:i:5:p:573-93}

\bibitem{hamsus}
Hamilton, J., Susmel, R.: Autoregressive conditional heteroskedasticity and
  changes in regime.
\newblock Journal of Econometrics \textbf{64}(1-2), 307--333 (1994).
\newblock
  \urlprefix\url{https://EconPapers.repec.org/RePEc:eee:econom:v:64:y:1994:i:1-2:p:307-333}

\bibitem{hamswar}
Hamilton, J.D.: {A} {N}ew {A}pproach to the {E}conomic {A}nalysis of
  {N}onstationary {T}ime {S}eries and the {B}usiness {C}ycle.
\newblock Econometrica \textbf{57}(2), 357--384 (1989).
\newblock \urlprefix\url{http://www.jstor.org/stable/1912559}

\bibitem{10.1016/j.elerap.2018.02.006}
Han, S., Hao, X., Huang, H.: An event-extraction approach for business analysis
  from online chinese news.
\newblock Electronic Commerce Research and Applications \textbf{28} (2018).
\newblock \doi{10.1016/j.elerap.2018.02.006}

\bibitem{DBLP:journals/corr/HillCK16}
Hill, F., Cho, K., Korhonen, A.: Learning distributed representations of
  sentences from unlabelled data.
\newblock CoRR \textbf{abs/1602.03483} (2016).
\newblock \urlprefix\url{http://arxiv.org/abs/1602.03483}

\bibitem{doi:10.1162/neco.1997.9.8.1735}
Hochreiter, S., Schmidhuber, J.: Long short-term memory.
\newblock Neural Computation \textbf{9}(8), 1735--1780 (1997).
\newblock \doi{10.1162/neco.1997.9.8.1735}.
\newblock \urlprefix\url{https://doi.org/10.1162/neco.1997.9.8.1735}

\bibitem{Hu:2018:LCW:3159652.3159690}
Hu, Z., Liu, W., Bian, J., Liu, X., Liu, T.Y.: Listening to chaotic whispers: A
  deep learning framework for news-oriented stock trend prediction.
\newblock In: Proceedings of the Eleventh ACM International Conference on Web
  Search and Data Mining, WSDM '18, pp. 261--269. ACM, New York, NY, USA
  (2018).
\newblock \doi{10.1145/3159652.3159690}.
\newblock \urlprefix\url{http://doi.acm.org/10.1145/3159652.3159690}

\bibitem{DBLP:journals/corr/LeM14}
Le, Q.V., Mikolov, T.: Distributed representations of sentences and documents.
\newblock CoRR \textbf{abs/1405.4053} (2014).
\newblock \urlprefix\url{http://arxiv.org/abs/1405.4053}

\bibitem{D14-1218}
Li, J., Hovy, E.: A model of coherence based on distributed sentence
  representation.
\newblock In: Proceedings of the 2014 Conference on Empirical Methods in
  Natural Language Processing (EMNLP), pp. 2039--2048. Association for
  Computational Linguistics, Doha, Qatar (2014).
\newblock \doi{10.3115/v1/D14-1218}.
\newblock \urlprefix\url{https://www.aclweb.org/anthology/D14-1218}

\bibitem{8068217}
{Li}, Q., {Chen}, Y., {Wang}, J., {Chen}, Y., {Chen}, H.: Web media and stock
  markets : A survey and future directions from a big data perspective.
\newblock IEEE Transactions on Knowledge and Data Engineering \textbf{30}(2),
  381--399 (2018).
\newblock \doi{10.1109/TKDE.2017.2763144}

\bibitem{The-effect-of-news-and-public-mood}
Li, Q., Wang, T., Li, P., Liu, L., Gong, Q., Chen, Y.: The effect of news and
  public mood on stock movements.
\newblock Information Sciences \textbf{278}, 826 -- 840 (2014).
\newblock \doi{10.1016/j.ins.2014.03.096}

\bibitem{DBLP:journals/corr/abs-1803-02893}
Logeswaran, L., Lee, H.: An efficient framework for learning sentence
  representations.
\newblock CoRR \textbf{abs/1803.02893} (2018).
\newblock \urlprefix\url{http://arxiv.org/abs/1803.02893}

\bibitem{long2019deep}
Long, W., Lu, Z., Cui, L.: Deep learning-based feature engineering for stock
  price movement prediction.
\newblock Knowledge-Based Systems \textbf{164}, 163--173 (2019)

\bibitem{long2019new}
Long, W., Song, L., Tian, Y.: A new graphic kernel method of stock price trend
  prediction based on financial news semantic and structural similarity.
\newblock Expert Systems with Applications \textbf{118}, 411--424 (2019)

\bibitem{doi:10.1080/14697688.2012.672762}
Luss, R., D'Aspremont, A.: Predicting abnormal returns from news using text
  classification.
\newblock Quantitative Finance \textbf{15}(6), 999--1012 (2015).
\newblock \doi{10.1080/14697688.2012.672762}.
\newblock \urlprefix\url{https://doi.org/10.1080/14697688.2012.672762}

\bibitem{DBLP:journals/corr/abs-1301-3781}
Mikolov, T., Chen, K., Corrado, G., Dean, J.: Efficient estimation of word
  representations in vector space.
\newblock CoRR \textbf{abs/1301.3781} (2013).
\newblock \urlprefix\url{http://arxiv.org/abs/1301.3781}

\bibitem{DBLP:journals/corr/MikolovSCCD13}
Mikolov, T., Sutskever, I., Chen, K., Corrado, G., Dean, J.: Distributed
  representations of words and phrases and their compositionality.
\newblock CoRR \textbf{abs/1310.4546} (2013).
\newblock \urlprefix\url{http://arxiv.org/abs/1310.4546}

\bibitem{Quantifying-Wikipedia-Usage}
Moat, H.S., Curme, C., Avakian, A., Kenett, D.Y., Stanley, H.E., Preis, T.:
  Quantifying wikipedia usage patterns before stock market moves.
\newblock Scientific Reports \textbf{3}, 1801 EP -- (2013).
\newblock \urlprefix\url{https://doi.org/10.1038/srep01801}

\bibitem{Perozzi:2014:DOL:2623330.2623732}
Perozzi, B., Al-Rfou, R., Skiena, S.: Deepwalk: Online learning of social
  representations.
\newblock In: Proceedings of the 20th ACM SIGKDD International Conference on
  Knowledge Discovery and Data Mining, KDD '14, pp. 701--710. ACM, New York,
  NY, USA (2014).
\newblock \doi{10.1145/2623330.2623732}.
\newblock \urlprefix\url{http://doi.acm.org/10.1145/2623330.2623732}

\bibitem{Rather:2015:RNN:2775746.2776067}
Rather, A.M., Agarwal, A., Sastry, V.: Recurrent neural network and a hybrid
  model for prediction of stock returns.
\newblock Expert Syst. Appl. \textbf{42}(6), 3234--3241 (2015).
\newblock \doi{10.1016/j.eswa.2014.12.003}.
\newblock \urlprefix\url{http://dx.doi.org/10.1016/j.eswa.2014.12.003}

\bibitem{azfin}
Schumaker, R., Chen, H.c.: Textual analysis of stock market prediction using
  breaking financial news: The azfin text system.
\newblock ACM Trans. Inf. Syst. \textbf{27} (2009).
\newblock \doi{10.1145/1462198.1462204}

\bibitem{e98ec1711a274278b32679ae3f451935}
Schumaker, R., Zhang, Y., Huang, C., Chen, H.: Evaluating sentiment in
  financial news articles.
\newblock Decision Support Systems \textbf{53}(3), 458--464 (2012).
\newblock \doi{10.1016/j.dss.2012.03.001}

\bibitem{Setty:2018:ENE:3209978.3210136}
Setty, V., Hose, K.: Event2vec: Neural embeddings for news events.
\newblock In: The 41st International ACM SIGIR Conference on Research \&\#38;
  Development in Information Retrieval, SIGIR '18, pp. 1013--1016. ACM, New
  York, NY, USA (2018).
\newblock \doi{10.1145/3209978.3210136}.
\newblock \urlprefix\url{http://doi.acm.org/10.1145/3209978.3210136}

\bibitem{article}
Shi, L., Teng, Z., Wang, L., Zhang, Y., Binder, A.: Deepclue: Visual
  interpretation of text-based deep stock prediction.
\newblock IEEE Transactions on Knowledge and Data Engineering \textbf{PP}, 1--1
  (2018).
\newblock \doi{10.1109/TKDE.2018.2854193}

\bibitem{si-etal-2014-exploiting}
Si, J., Mukherjee, A., Liu, B., Pan, S.J., Li, Q., Li, H.: Exploiting social
  relations and sentiment for stock prediction.
\newblock In: Proceedings of the 2014 Conference on Empirical Methods in
  Natural Language Processing ({EMNLP}), pp. 1139--1145. Association for
  Computational Linguistics, Doha, Qatar (2014).
\newblock \doi{10.3115/v1/D14-1120}.
\newblock \urlprefix\url{https://www.aclweb.org/anthology/D14-1120}

\bibitem{DBLP:journals/corr/VaswaniSPUJGKP17}
Vaswani, A., Shazeer, N., Parmar, N., Uszkoreit, J., Jones, L., Gomez, A.N.,
  Kaiser, L., Polosukhin, I.: Attention is all you need.
\newblock CoRR \textbf{abs/1706.03762} (2017).
\newblock \urlprefix\url{http://arxiv.org/abs/1706.03762}

\bibitem{Vincent:2010:SDA:1756006.1953039}
Vincent, P., Larochelle, H., Lajoie, I., Bengio, Y., Manzagol, P.A.: Stacked
  denoising autoencoders: Learning useful representations in a deep network
  with a local denoising criterion.
\newblock J. Mach. Learn. Res. \textbf{11}, 3371--3408 (2010).
\newblock \urlprefix\url{http://dl.acm.org/citation.cfm?id=1756006.1953039}

\bibitem{Wang:2012:NTM:2142138.2142443}
Wang, B., Huang, H., Wang, X.: A novel text mining approach to financial time
  series forecasting.
\newblock Neurocomput. \textbf{83}, 136--145 (2012).
\newblock \doi{10.1016/j.neucom.2011.12.013}.
\newblock \urlprefix\url{http://dx.doi.org/10.1016/j.neucom.2011.12.013}

\bibitem{wang2019integrated}
Wang, P., Zong, L., Ma, Y.: An integrated early warning system for stock market
  turbulence.
\newblock arXiv preprint arXiv:1911.12596  (2019)

\bibitem{725072}
{Wuthrich}, B., {Cho}, V., {Leung}, S., {Permunetilleke}, D., {Sankaran}, K.,
  {Zhang}, J.: Daily stock market forecast from textual web data.
\newblock In: SMC'98 Conference Proceedings. 1998 IEEE International Conference
  on Systems, Man, and Cybernetics (Cat. No.98CH36218), vol.~3, pp. 2720--2725
  vol.3 (1998).
\newblock \doi{10.1109/ICSMC.1998.725072}

\bibitem{N16-1174}
Yang, Z., Yang, D., Dyer, C., He, X., Smola, A., Hovy, E.: Hierarchical
  attention networks for document classification.
\newblock In: Proceedings of the 2016 Conference of the North American Chapter
  of the Association for Computational Linguistics: Human Language
  Technologies, pp. 1480--1489. Association for Computational Linguistics, San
  Diego, California (2016).
\newblock \doi{10.18653/v1/N16-1174}.
\newblock \urlprefix\url{https://www.aclweb.org/anthology/N16-1174}

\bibitem{8759115}
{Zheng}, J., {Xia}, A., {Shao}, L., {Wan}, T., {Qin}, Z.: Stock volatility
  prediction based on self-attention networks with social information.
\newblock In: 2019 IEEE Conference on Computational Intelligence for Financial
  Engineering Economics (CIFEr), pp. 1--7 (2019).
\newblock \doi{10.1109/CIFEr.2019.8759115}

\end{thebibliography}

\end{document}